# Subgraph Matching Kernels for Attributed Graphs

Nils Kriege                                    NILS.KRIEGE@CS.TU-DORTMUND.DE
Petra Mutzel                                   PETRA.MUTZEL@CS.TU-DORTMUND.DE
Department of Computer Science, Technische Universität Dortmund, Germany

## Abstract

We propose graph kernels based on subgraph matchings, i.e. structure-preserving bijections between subgraphs. While recently proposed kernels based on common subgraphs (Wale et al., 2008; Shervashidze et al., 2009) in general can not be applied to attributed graphs, our approach allows to rate mappings of subgraphs by a flexible scoring scheme comparing vertex and edge attributes by kernels. We show that subgraph matching kernels generalize several known kernels. To compute the kernel we propose a graph-theoretical algorithm inspired by a classical relation between common subgraphs of two graphs and cliques in their product graph observed by Levi (1973). Encouraging experimental results on a classification task of real-world graphs are presented.

## 1. Introduction

Graphs are well-studied versatile representations of structured data and have become ubiquitous in many application domains like chem- and bioinformatics. Comparing graphs is a fundamental problem and computing meaningful similarity measures is a prerequisite to apply a variety of machine learning algorithms to the domain of graphs. Consequently related problems have been extensively studied involving essential graph theoretical questions, which are typically $\mathcal{NP}$-hard, like, e.g, the maximum common subgraph problem. However, graph similarity can be defined in various ways and its computation not necessarily requires to solve these problems exactly to yield a similarity measure appropriate for a wide range of applications.

To become applicable to the wealth of so-called *ker-*

nel methods, including Support Vector Machines as the most prominent example, similarity measures must satisfy the additional constraints to be symmetric and positive semidefinite (p.s.d.). While recent development of graph kernels primarily focuses on large datasets of graphs with simple labels (cf. Table 1), it has been observed on several occasions that the prediction accuracy can be increased by annotating vertices or edges with additional attributes (see, e.g., Borgwardt et al., 2005; Fröhlich et al., 2005; Harchaoui & Bach, 2007). Since attributes in many cases include continuous values, a meaningful similarity measure must tolerate certain divergence. Therefore, kernels designed for graphs with simple labels often are not suitable for attributed graphs.

We propose a new graph kernel which is related to the maximum common subgraph problem. Instead of deriving a similarity measure from a maximum common subgraph our approach counts the number of matchings between subgraphs up to a fixed size and therefore has polynomial runtime. Attributes of mapped vertices and edges are assessed by a flexible scoring scheme and, thus, the approach can be applied to general attributed graphs.

### 1.1. Related Work

In recent years various graph kernels have been proposed, see (Vishwanathan et al., 2010) and references therein. Gärtner et al. (2003) and Kashima et al. (2003) devised graph kernels based on random walks, which count the number of labeled walks two graphs have in common. Subsequently random walk kernels were extended to avoid repeated consecutive vertices and were combined with vertex label enrichment techniques by Mahé et al. (2004). The runtime was improved particularly for graphs with simple labels (Vishwanathan et al., 2010). Random walk kernels have been extended to take vertex and edge attributes into account and were thereby successfully applied to protein function prediction (Borgwardt et al., 2005).

A drawback of random walks is that walks are struc-turally simple. However, computing kernels by counting common subgraphs of unbounded size is known to be $\mathcal{NP}$-complete (Gärtner et al., 2003). Thus, another direction in the development of graph kernels focuses on small subgraphs of a fixed size $k \in \{3, 4, 5\}$, referred to as *graphlets*, which primarily apply to unlabeled graphs (Shervashidze et al., 2009). Furthermore tree patters, which are allowed to contain repeated vertices just like random walks, were proposed by Ramon & Gärtner (2003) and later refined by Mahé & Vert (2009). While these approaches are based on all common subtree patterns of a specified height, others only take the entire neighborhood of each vertex up to given distance into account (Shervashidze et al., 2011), thus reducing the number of features and the required runtime significantly. Menchetti et al. (2005) proposed a weighted decomposition kernel, which determines matching substructures by a restrictive kernel (*selector*) and weights each matching by a kernel defined on the context of the matching. A kernel based on shortest-paths was developed by Borgwardt & Kriegel (2005), which first computes the length of shortest-paths between all pairs of vertices and then counts pairs with similar labels and distance. Instead of comparing pairs of individual vertices, the kernel proposed by Costa & De Grave (2010) associates a string encoding the neighborhood subgraph with each vertex.

*Table 1.* Summary on (selected) graph kernels regarding computation by explicit feature mapping (EXP. $\phi$) and support for attributed graphs (ATTR.).

| GRAPH KERNEL | EXP. $\phi$ | ATTR. |
|---|---|---|
| RW (Gärtner et al., 2003) | $\times$ | $\checkmark$ |
| TP (Ramon & Gärtner, 2003) | $\times$ | $\checkmark$ |
| SP (Borgwardt & Kriegel, 2005) | $\times$ | $(\checkmark)$ |
| Graphlet (Shervashidze et al., 2009) | $\checkmark$ | $\times$ |
| NSPDK (Costa & De Grave, 2010) | $\checkmark$ | $\times$ |
| WL (Shervashidze et al., 2011) | $\checkmark$ | $\times$ |

Several graph kernels were tailored especially to chemical compound. For attributed molecular graphs Fröhlich et al. (2005) proposed a similarity measure based on an optimal assignment of vertices. However, the proposed function was shown not to be p.s.d. (Vishwanathan et al., 2010). Established techniques in cheminformatics are based on features which can be 1. directly generated from the molecular graph, e.g. all paths or subgraphs up to a certain size (Wale et al., 2008), similar to graphlets, 2. taken from a predefined dictionary or 3. generated in a preceding data-mining phase, e.g. using frequent subgraph mining.

The proposed techniques can be classified into approaches that use explicit feature mapping and those that directly compute a kernel function. If explicit representations are manageable, these approaches usually outperform other kernels regarding runtime on large datasets, since the number of vector representations scales linear with the dataset size. However, these approaches do not support attributed graphs, cf. Table 1. The computation technique proposed for random walk and tree pattern kernels, in contrast, can be extended to compare vertex and edge attributes by kernels. However, compared to graphlet kernels these approaches are based on simple features including repeated vertices.

We propose a technique that uses small subgraphs contained in the two graphs under comparison, similar to graphlets, but simultaneously provide the flexibility to compare vertex and edge attributes by means of arbitrary kernel functions.

## 2. Preliminaries

In this section basic concepts of graph theory are introduced. We refer to simple undirected graphs. Given a graph $G = (V, E)$ we denote by $V(G) = V$ and $E(G) = E$ the set of *vertices* and *edges*, respectively. The set of vertices *adjacent* to a vertex $v$ is denoted by $N(v) = \{u \in V : (u, v) \in E\}$. A *path* of length $n$ is a sequence of vertices $(v_0, \ldots, v_n)$ such that $(v_i, v_{i+1}) \in E$ for $0 \le i < n$. A graph is *connected* if at least one path between any pair of vertices exists and *disconnected* otherwise. A graph $G' = (V', E')$ is a *subgraph* of a graph $G = (V, E)$, written $G' \subseteq G$, iff $V' \subseteq V$ and $E' \subseteq E$. If $E' = (V' \times V') \cap E$ holds, $G' = G[V']$ is said to be *induced* by $V'$ in $G$. Note that a subgraph of a connected graph may be disconnected. In the following we will always refer to induced subgraphs and assume graphs to be *labeled* or *attributed*, i.e. a graph is a 3-tuple $G = (V, E, l)$, where $l : V \cup E \to \mathcal{L}$ is a *labeling function* associating the label $l(v)$ to the vertex $v$ and $l(e)$ to the edge $e$. All labels are from the set $\mathcal{L}$ and may as well consist of tuples of attribute-value pairs.

A *graph isomorphism* between two labeled graphs $G_1 = (V_1, E_1, l_1)$ and $G_2 = (V_2, E_2, l_2)$ is a bijection $\varphi : V_1 \to V_2$ that preserves adjacencies, i.e. $\forall u, v \in V_1 : (u, v) \in E_1 \Leftrightarrow (\varphi(u), \varphi(v)) \in E_2$, and labels: Let $\psi_\varphi : V_1 \times V_1 \to V_2 \times V_2$ be the mapping of vertex pairs implicated by the bijection $\varphi$ such that $\psi_\varphi((u, v)) = (\varphi(u), \varphi(v))$. Then to preserve labels the conditions $\forall v \in V_1 : l_1(v) \equiv l_2(\varphi(v))$ and $\forall e \in E_1 : l_1(e) \equiv l_2(\psi_\varphi(e))$ must hold, where $\equiv$ denotes that two labels are considered equivalent. Two graphs $G_1$, $G_2$ are said to be *isomorphic*, written $G_1 \simeq G_2$, if a graph isomorphism between $G_1$ and $G_2$ exists. An *automorphism* of a graph $G = (V, E)$ is a graph isomorphism $\varphi : V \to V$. The set of automorphisms of $G$ is denoted by $\mathrm{Aut}(G)$.

## 3. A Subgraph Matching Kernel

Several graph kernels count the number of isomorphic subgraphs contained in two graphs. A common subgraph isomorphism in contrast denotes a mapping between such subgraphs that preserves their structure.

**Definition 1 (Common Subgraph Isomorphism)**
Let $G_1 = (V_1, E_1, l_1)$, $G_2 = (V_2, E_2, l_2)$ be two graphs and $V_1' \subseteq V_1$, $V_2' \subseteq V_2$ subsets of their vertices. A graph isomorphism $\varphi$ of $G_1[V_1']$ and $G_2[V_2']$ is called *common subgraph isomorphism* (CSI) of $G_1$ and $G_2$. □

Based on this definition we define the following kernel and will see later that the function is p.s.d.

**Definition 2 (CSI Kernel)** Let $\mathcal{I}(G_1, G_2)$ denote the set of all CSIs of two graphs $G_1$ and $G_2$ and $\lambda : \mathcal{I}(G_1, G_2) \to \mathbb{R}^+$ a weight function. The function

$$k_{\mathrm{csi}}(G_1, G_2) = \sum_{\varphi \in \mathcal{I}(G_1, G_2)} \lambda(\varphi) \qquad (1)$$

is called *common subgraph isomorphism kernel*. □

When vertices and edges are annotated with arbitrary attributes it is inappropriate to require a mapping to preserve the structure and the labels of the two graphs exactly. To this end, we generalize Def. 2 to allow for a more flexible scoring of bijections referred to as *graph matching*.

**Definition 3 (Subgraph Matching Kernel)**
Given two graphs $G_1 = (V_1, E_1, l_1)$, $G_2 = (V_2, E_2, l_2)$, let $\mathcal{B}(G_1, G_2)$ denote the set of all bijections between sets $V_1' \subseteq V_1$ and $V_2' \subseteq V_2$ and let $\lambda : \mathcal{B}(G_1, G_2) \to \mathbb{R}^+$ be a weight function. The *subgraph matching kernel* is defined as

$$k_{\mathrm{sm}}(G_1, G_2) = \sum_{\varphi \in \mathcal{B}(G_1, G_2)} \lambda(\varphi) \prod_{v \in V_1'} \kappa_V(v, \varphi(v)) \prod_{e \in V_1' \times V_1'} \kappa_E(e, \psi_\varphi(e)),$$

where $V_1' = \mathrm{dom}(\varphi)$ and $\kappa_V$, $\kappa_E$ kernel function defined on vertices and pairs of vertices, respectively. □

**Theorem 1** *The subgraph matching kernel is p.s.d.* □

PROOF The structure of a graph $G = (V, E)$ with $n$ vertices can be encoded by a tuple $(\vec{v}, \mathbf{e})$, where $\vec{v} = (v_i)_n$ is a sequence of the vertices in $V$ and $\mathbf{e} = [e_{i,j}]_{n \times n}$ is a matrix of elements $E \cup \{\epsilon\}$, such that $e_{i,j} = (v_i, v_j)$ if $(v_i, v_j) \in E$ and $\epsilon$ otherwise. By extending $\vec{v}$ and $\mathbf{e}$ by additional $\epsilon$-elements we can encode graphs of different size into the same space. Each permutation of the vertices of a graph yields a valid encoding and a graph can be decomposed into all its encodings. This allows us to define a graph kernel by specifying an $R$-convolution (Haussler, 1999). Let $R(\vec{v}, \mathbf{e}, G)$ be a relation, where $\vec{v}$ and $\mathbf{e}$ are defined as above, $G$ is a graph and $R(\vec{v}, \mathbf{e}, G) = 1$ iff $(\vec{v}, \mathbf{e})$ is an encoding of $G$. Let $R^{-1}(G) = \{(\vec{v}, \mathbf{e}) : R(\vec{v}, \mathbf{e}, G) = 1\}$ be the set of encodings of $G$. We can now specify the $R$-convolution kernel

$$k_{\mathrm{enc}}(G_1, G_2) = \sum_{\substack{(\vec{v}, \mathbf{e}) \in R^{-1}(G_1) \\ (\vec{v}, \mathbf{e}) \in R^{-1}(G_2)}} \prod_i \kappa_V(u_i, v_i) \prod_{i,j} \kappa_E(e_{i,j}, f_{i,j}),$$

where $\kappa_V(\epsilon, \cdot) = 0$. Combining this kernel with a convolution kernel based on subgraph decomposition and a suitable weight function yields

$$k(G_1, G_2) = \sum_{G_1' \subseteq G_1} \sum_{G_2' \subseteq G_2} \frac{1}{|V(G_1')|!} k_{\mathrm{enc}}(G_1', G_2'). \quad (2)$$

This kernel is equivalent to $k_{\mathrm{sm}}$ with $\lambda(\varphi) = 1$, since there are exactly $n!$ pairs of encodings of two graphs with $n$ vertices corresponding to the same bijection. ∎

We can identify Def. 2 as a special case of Def. 3, where

$$\kappa_V(v_1, v_2) = \begin{cases} 1 & \text{if } l_1(v_1) \equiv l_2(v_2), \\ 0 & \text{otherwise and} \end{cases}$$

$$\kappa_E(e_1, e_2) = \begin{cases} 1 & \text{if } e_1 \in E_1 \wedge e_2 \in E_2 \wedge l_1(e) \equiv l_2(e) \\ & \text{or } e_1 \notin E_1 \wedge e_2 \notin E_2, \\ 0 & \text{otherwise.} \end{cases}$$

These kernels assure that exactly the conditions of graph isomorphism are fulfilled. Therefore the CSI kernel is a special case of the subgraph matching kernel and we may state the following corollary.

**Corollary 1** *The CSI kernel is p.s.d.* □

### 3.1. Relation to the Subgraph Kernel

The definitions of subgraph kernels proposed slightly differs. Here we refer to induced subgraphs of unbounded size.

**Definition 4 (Subgraph Kernel)** Given two graphs $G_1, G_2 \in \mathcal{G}$ and a weight function $\lambda_{\mathrm{s}} : \mathcal{G} \to \mathbb{R}^+$. The *subgraph kernel* is defined as

$$k_{\mathrm{s}}(G_1, G_2) = \sum_{G_1' \subseteq G_1} \sum_{G_2' \subseteq G_2} \lambda_{\mathrm{s}}(G_1') k_{\simeq}(G_1', G_2'), \quad (3)$$

where $k_{\simeq} : \mathcal{G} \times \mathcal{G} \to \{0, 1\}$ is the isomorphism kernel, i.e. $k_{\simeq}(G_1', G_2') = 1$ iff $G_1' \simeq G_2'$. □

The subgraph kernel basically counts isomorphic subgraphs, while the CSI kernel counts the number of isomorphisms between subgraphs. Since there may be more than one isomorphism between a pair of isomorphic subgraphs, both concepts differ in detail.

**Theorem 2** *Let $k_s$ be a subgraph kernel with weight function $\lambda_s$ and $k_{csi}$ a CSI kernel with weight function $\lambda_{csi}(\varphi) = \frac{\lambda_s(G)}{|\operatorname{Aut}(G)|}$, where $G = G_1[\operatorname{dom}(\varphi)]$. Then $k_{csi}(G_1, G_2) = k_s(G_1, G_2)$ for all graphs $G_1, G_2 \in \mathcal{G}$.* □

PROOF For each pair $(G_1', G_2')$ that contributes to the sum of Eq. (3), $G_1' \simeq G_2'$ holds. CSIs exist for these pairs of graphs only. There are $|\operatorname{Aut}(G_1')| = |\operatorname{Aut}(G_2')|$ isomorphism between $G_1'$ and $G_2'$, each of which is contained in $\mathcal{I}(G_1, G_2)$ and contributes to Eq. (1). This is compensated by the correction term $|\operatorname{Aut}(G_1')|^{-1}$.■

### 3.2. Relation to the Pharmacophore Kernel

Mahé et al. (2006) proposed a kernel to compare chemical compounds based on characteristic features together with their relative spatial arrangement, so-called *pharmacophores*. To this end, a molecule is represented by a set of pairs $M = \{(x_i, l_i) \in \mathbb{R}^3 \times \mathcal{L}\}_i$, where $x_i$ are the coordinates of a feature $i$ in a 3-dimensional space and $l_i$ is an associated label. The set of pharmacophores of a molecule $M$ is $\mathcal{P}(M) = \{(a_1, a_2, a_3) \in M^3 : a_1 \neq a_2, a_1 \neq a_3, a_2 \neq a_3\}$. The pharmacophore kernel is then defined as

$$k_p(M_1, M_2) = \sum_{p_1 \in \mathcal{P}(M_1)} \sum_{p_2 \in \mathcal{P}(M_2)} k_i(p_1, p_2) k_s(p_1, p_2)$$

and measures the similarity of two molecules based on triples of similar characteristic features with a similar spatial arrangement, which is quantified by the two kernels $k_i$ and $k_s$, respectively. These are defined as $k_i(p, p') = \prod_{i=1}^{3} k_{\text{feat}}(l_i, l_i')$ and $k_s(p, p') = \prod_{i=1}^{3} k_{\text{dist}}(\|x_i, x_{i+1}\|, \|x_i', x_{i+1}'\|)$, where $\|\cdot\|$ denotes the Euclidean distance and the index $i+1$ is taken modulo 3.

From the representation $M$ of a molecule as used by the pharmacophore kernel we can construct a graph $G(M) = (V_M, E_M, l_M)$, such that $V_M = \{v_1, \ldots, v_{|M|}\}$ with $l_M(v_i) = l_i$ and $E_M = V_M \times V_M$ with $l_M((v_i, v_j)) = \|x_i, x_j\|$.

**Theorem 3** *Let $k_p$ be a pharmacophore kernel and $k_{sm}$ a subgraph matching kernel with weight function $\lambda(\varphi) = 6$ if $|\operatorname{dom}(\varphi)| = 3$ and 0 otherwise and vertex and edge kernels defined as $\kappa_V(v_1, v_2) = k_{\text{feat}}(l(v_1), l(v_2))$ and $\kappa_E(e_1, e_2) = k_{\text{dist}}(l(e_1), l(e_2))$. Then $k_p(M_1, M_2) = k_{sm}(G(M_1), G(M_2))$ holds.* □

PROOF The weight function $\lambda$ ensures that only subgraphs with three vertices contribute to the value of $k_{sm}$. Since $G(M)$ is a complete graph by definition, each common subgraph induced by three vertices is a triangle, i.e. all triples of vertices with their pairwise distances are taken into account. For each subset with three vertices there are six different triples representing all possible permutations. Comparing two subsets with three elements, $3! = 6$ combinations of associated triples correspond to the same mapping of three vertices. Thus, multiplying the value of 3-element subgraph matchings by 6 compensates for this. ■

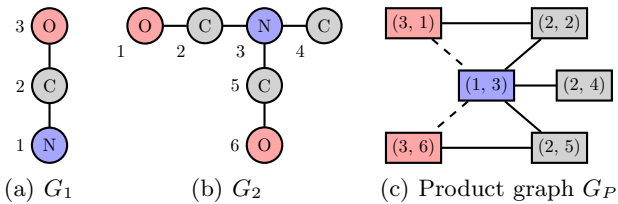

*Figure 1.* Two labeled graphs (a), (b) and their product graph $G_P$ (c); dashed lines represent d-edges.

### 3.3. Kernel Computation

In this section we propose an algorithm to compute the CSI and subgraph matching kernel. Our technique is inspired by a classical result of Levi (1973) who observed a relation between common subgraphs of two graphs and cliques in their product graph. Given two graphs $G_1 = (V_1, E_1, l_1)$ and $G_2 = (V_2, E_2, l_2)$, the *(modular) product graph* $G_P = (V_P, E_P)$ of $G_1$ and $G_2$ is defined by $V_P = \{(v_1, v_2) \in V_1 \times V_2 : l_1(v_1) \equiv l_2(v_2)\}$ and $E_P$ containing an edge connecting two vertices $(u_1, u_2), (v_1, v_2) \in V_P$ iff $u_1 \neq v_1, u_2 \neq v_2$ and either $e_1 = (u_1, v_1) \in E_1$ and $e_2 = (u_2, v_2) \in E_2$ and $l_1(e_1) \equiv l_2(e_2)$ (*c-edge*) or $e_1 \notin E_1$ and $e_2 \notin E_2$ (*d-edge*). The distinction of c-edges and d-edges is due to Koch (2001); c-edges represent common adjacency and d-edges common non-adjacency[1]. Thus, two vertices of the product graph are adjacent iff the corresponding vertex mappings can be part of the same CSI, see Fig. 1 for an example.

Levi (1973) observed that each maximum clique in the product graph is associated with a maximum common subgraph of the factor graphs. Furthermore, the vertex set $C$ is a clique in $G_P$ iff there is a corresponding CSI $\varphi \in \mathcal{I}(G_1, G_2)$. As a consequence we can enumerate (or count) all CSIs by enumerating (counting) the cliques of the product graph. To compute the subgraph matching kernel we extend the approach by means of a weighted product graph, where vertices and edges are annotated with the values of $\kappa_V$ and $\kappa_E$, respectively. Each clique is then associated with the product of the weights of all vertices and edges contained in it.

---

[1] The concept of product graphs has been used for the computation of graph kernels before. Note that the definition used here is different from the direct product graph, which contains only c-edges, proposed by Gärtner et al. (2003) to compute a random walk kernel.

**Definition 5 (Weighted Product Graph)** Given two graphs $G_1 = (V_1, E_1, l_1)$, $G_2 = (V_2, E_2, l_2)$ and vertex and edge kernels $\kappa_V$ and $\kappa_E$, the *weighted product graph* (WPG) $G_P = (V_P, E_P, c)$ of $G_1$ and $G_2$ is defined by

$$
\begin{aligned}
V_P &= \{(v_1, v_2) \in V_1 \times V_2 : \kappa_V(v_1, v_2) > 0\} \\
E_P &= \{((u_1, u_2), (v_1, v_2)) \in V_P \times V_P : u_1 \neq v_1 \wedge \\
&\qquad u_2 \neq v_2 \wedge \kappa_E((u_1, v_1), (u_2, v_2)) > 0\} \\
c(v) &= \kappa_V(v_1, v_2) \qquad \forall v = (v_1, v_2) \in V_P \\
c(e) &= \kappa_E((u_1, v_1), (u_2, v_2)) \qquad \forall e \in E_P,
\end{aligned}
$$

where $e = ((u_1, u_2), (v_1, v_2))$. □

If we assume $\kappa_E(e_1, e_2) = 0$ if $e_1 \in E_1$ and $e_2 \notin E_2$ or vice versa, the distinction of c- and d-edges carries over to weighted product graphs.

---

**Algorithm 1:** SMKERNEL($w$, $C$, $P$)

**Input** : WPG $G_P = (V_P, E_P, c)$, weight function $\lambda$
**Initial** : $value \leftarrow 0$; SMKERNEL($1, \emptyset, V_P$)
**Param.**: Weight $w$ of the clique $C$, candidate set $P$
**Output**: Result of the kernel function $value$

1 **while** $|P| > 0$ **do**
2     $v \leftarrow$ arbitrary element of $P$
3     $C' \leftarrow C \cup \{v\}$
4     $w' \leftarrow w \cdot c(v)$     ▷ *multiply by vertex weight*
5     **forall** the $u \in C$ **do**
6        $w' \leftarrow w' \cdot c(u, v)$     ▷ *multiply by edge weights*
7     $value \leftarrow value + w' \cdot \lambda(C')$
8     SMKERNEL($w'$, $C'$, $P \cap N(v)$)     ▷ *extend clique*
9     $P \leftarrow P \setminus \{v\}$

---

Algorithm 1 computes the subgraph matching kernel by enumeration of cliques. A current clique is extended stepwise by all vertices preserving the clique property. These vertices form the candidate set $P$. Whenever the current clique $C$ is extended by a new vertex $v$, the weight of the vertex itself (line 4) and all the edges connecting $v$ to a vertex in $C$ (line 6) are multiplied with the weight of the current clique $w$ to obtain the weight of the new clique. The algorithm effectively avoids duplicates by removing a vertex from the candidate set after all cliques containing it have been exhaustively explored (line 9).

### 3.3.1. RESTRICTION TO SUBGRAPH CLASSES

In this section we discuss restrictions to certain classes of subgraphs, their relation to cliques in the product graph and appropriate modifications of the enumeration algorithm. Since finding a maximum clique or a maximum CSI is known to be an $\mathcal{NP}$-hard problem, it may be required in practice to restrict the size of the subgraphs considered. Modifying Algorithm 1 to stop the recursion whenever a fixed maximum size $k$ has been reached, effectively restricts the size of the cliques and thereby the size of the matched subgraphs, which is quantified by the number of vertices.

Restricting to connected subgraphs may also significantly reduce the search space, especially when graphs are sparse. Moreover disconnected subgraphs convey less structural information and may therefore be considered less relevant. This constraint can be realized by an adaption of a technique proposed by Koch (2001). A clique that is spanned by c-edges, a so-called *c-clique*, corresponds to a connected CSI. Algorithm 1 can be modified to only enumerate c-cliques by making sure that only vertices are added that are adjacent to a vertex in the current clique via at least one c-edge. The restricted variants remain p.s.d., since they are equivalent to the general subgraph matching kernel with a suitably chosen weight function.

### 3.3.2. RUNTIME ANALYSIS

**Complexity** The runtime of Algorithm 1 depends on the number of cliques in the product graph. Since there is a one-to-one correspondence between cliques and bijections contributing to the kernel value, we can derive an upper bound for the number of cliques in $G_P$ by considering the number of possible bijections. There are $\binom{n}{k}$ induced subgraphs of size $k$ in a graph with $n$ vertices and up to $k!$ isomorphisms between graphs of size $k$. Thus, we have

$$
C(k) = \sum_{i=0}^{k} i! \binom{n_1}{i} \binom{n_2}{i} \leq \sum_{i=0}^{k} \binom{n_1 n_2}{i}
$$

cliques of size up to $k$ in $G_P$. Therefore the worst-case runtime of Algorithm 1 (modified to stop recursion at depth $k$) is $\mathcal{O}(nC(k)) = \mathcal{O}(kn^{k+1})$, where $n = n_1 n_2$.

**Practical considerations** The analysis of the complexity shows that a reasonable performance in practice can only be expected when the maximum size of the subgraphs considered is restricted. Therefore, the approach competes against subgraph or graphlet kernels. Besides the differences described in Sec. 3.1, the methods of computation exhibit substantially different characteristics: The runtime of our algorithm heavily depends on the number of allowed mappings of subgraphs. For instances with diverse labels in combination with a restrictive vertex kernel (e.g. Dirac kernel) the size of the product graph is typically substantially reduced, such that $|V_P| \ll |V_1| \cdot |V_2|$. In a similar way diverse edge labels may diminish the number of edges. Due to d-edges sparse graphs tend to have dense product graphs and contain a large number of cliques. However, in this case the number of enumerated cliques can be significantly reduced by restricting to c-cliques.

The computation of subgraph kernels is based on explicit mapping into feature space. While this is beneficial for certain datasets, the number of subgraphs quickly becomes very large for graphs with diverse labels rendering explicit mapping prohibitive (Shervashidze et al., 2009). Furthermore, subgraph kernels are not applicable to attributed graphs. In these respects, our approach is complementary to subgraph kernels and shows promise for instances for which these approaches fall short.

## 4. Experimental Evaluation

### 4.1. Method & Datasets

We performed classification experiments using the $C$-SVM implementation LIBSVM[2]. We report mean prediction accuracies as well as standard deviations obtained by 10-fold cross-validation repeated 10 times with random fold assignment. Within each fold the parameter $C$ was chosen from $\{10^{-3}, 10^{-2}, \ldots, 10^{3}\}$ by cross-validation based on the training set.

We compared the subgraph matching kernel (SM and CSM with connection constraint) to kernels based on fixed length random walks (FLRW) and tree patterns[3] (TP), both supporting attributed graphs. Our implementation is similar to the efficient dynamic programming approach proposed by Harchaoui & Bach (2007). Furthermore, we compare to the Geometric Random Walk (GRW), Shortest Path (SP), Weisfeiler-Lehman Subtree (WL) and Weisfeiler-Lehman Shortest Path (WLSP) kernel. WLSP is similar to NSPDK recently proposed by Costa & De Grave (2010).

These graph kernels can be tuned by several parameters. The maximum size of (C)SM was chosen from $k \in \{1, 2, \ldots, 7\}$ and a uniform weight function was used. FLRW was computed for walks of length $l \in \{1, 2, \ldots, 8\}$ and the parameter $\lambda$ for GRW was cho-sen from $\{10^{-5}, 10^{-4}, \ldots, 10^{-2}\}$. For TP we used a uniform weight $\lambda$ chosen from $\{10^{-5}, 10^{-4}, \ldots, 10^{-2}\}$ with height $h \in \{1, 2, 3, 4\}$. The number of iterations of WL/WLSP was chosen from $h \in \{0, 1, \ldots, 5\}$. All parameters were selected by cross-validation on the training datasets only. As remarked before (Wale et al., 2008; Costa & De Grave, 2010) kernels using features of different size are typically biased towards large features. Therefore, we also normalized kernel values separately for each feature size where applicable. Since the runtimes may depend on the selected parameters, we report the time required to compute a complete Gram matrix for each dataset using parameters frequently selected by the optimization process.

For a fair comparison all kernels were adapted to take vertex and edge labels into account and implemented in Java. For the pharmacophore kernel (PH) we used the implementation provided by the authors[4]. Experiments were conducted using Sun Java JDK v1.6.0 on an Intel Xeon E5430 machine at 2.66GHz with 8GB of RAM using a single processor only.

**Graphs with simple labels** We employed benchmark datasets containing molecules[5] and proteins: The MUTAG dataset consists of 188 chemical compounds divided into two classes according to their mutagenic effect on a bacterium. The PTC dataset contains compounds labeled according to carcinogenicity on rodents divided into male mice (MM), male rats (MR), female mice (FM) and female rats (FR). Molecules can naturally be represented by graphs, where vertices represent atoms and edges represent chemical bonds. We have removed explicit hydrogen atoms and labeled vertices by atom type and edges by bond type (single, double, triple or aromatic).

We have obtained the dataset ENZYME from Borgwardt et al. (2005), which is associated with the task of assigning 600 enzymes to one of the 6 EC top level classes. Vertices represent secondary structure elements (SSE) and are annotated by their type, i.e. helix/sheet/turn. Two vertices are connected by an edge if they are neighbors along the amino acid sequence or one of three nearest neighbors in space. Edges are annotated with their type, i.e. structural/sequential.

**Attributed graphs** Benchmark datasets containing attributed graphs are less wide-spread. We used the ENZYME dataset adding an attribute representing the 3d length of the SSE in Å to each vertex. The vertex kernel was defined as the product of a Dirac kernel on the type attributes and the Brownian bridge kernel with parameter $c = 3$ originally used on the length attribute, see (Borgwardt et al., 2005). The edge kernel remains a Dirac kernel on the type attribute.

---

[2] `http://www.csie.ntu.edu.tw/~cjlin/libsvm`

[3] The definition by Ramon & Gärtner (2003) is vague regarding subtrees without children. We require subtrees to have at least one child, see (Mahé & Vert, 2009) for a detailed discussion of this issue.

[4] ChemCPP v1.0.2, `http://chemcpp.sourceforge.net`

[5] Both datasets are widely used (see, e.g., Kashima et al., 2003) and can be obtained from `http://cdb.ics.uci.edu`

Further classification problems were derived from the chemical compound datasets BZR, COX2, DHFR and ER which come with 3D coordinates, and were used by Mahé et al. (2006) to study the pharmacophore kernel. We generated complete graphs from the compounds, where edges are labeled with distances[6] as described in Sect. 3.2 and vertex labels correspond to atom types. We used a triangular kernel to compare distances defined by $k(d_1, d_2) = \frac{1}{c} \cdot \max\{0, c - |d_1 - d_2|\}$ and chose $c$ from $\{0.1, 0.25, 0.5, 1.0\}$ by cross-validation.

## 4.2. Results & Discussion

The classification accuracies and runtimes are summarized in Tables 2 and 3. In terms of classification accuracy on graphs with simple labels no general suggestion which kernel performs best can be derived. CSM performs best on FM, where walk-based kernels perform slightly worse. For the multiclass classification problem ENZYME with simple labels CSM yields results comparable to WL and WLSP, while others perform worse. This observation also holds for ENZYME with attributes, where WL and WLSP can no longer be applied. All approaches benefit significantly from the additional vertex annotations, which indicates the importance of attributes, and CSM reaches the highest classification accuracy. On molecular distance graphs we observed that SM performs best in two of four cases and competitive on the other datasets. However, the differences here are rather small. Mahé et al. (2006) suggested to extend the pharmacophore kernel to take more than 3 points into account. At least for the instances we have tested, we observed that this does not lead to a significant increase in classification accuracy. Nevertheless, this might prove useful where more complex binding mechanism must be considered.

The runtime results on graphs with simple labels clearly show that computation schemes based on explicit mapping outperform other approaches. These all lie in the same order of magnitude and CSM is slower than FLRW and TP. For attributed graphs SP no longer allows explicit mapping. This leads to a considerable increase in runtime of SP on the ENZYME dataset, where it is now the slowest of the four tested approaches, while the other kernels noticeably benefit from the sparsity introduced by the vertex kernel taking the length attribute into account. TP could not be employed to molecular distance graphs, since the runtime to compute a Gram matrix here exceeded 24h even for the most restrictive edge kernel with $c = 0.1$. This can be explained by the fact that this class of graphs contains vertices with large sets of matching neighbors, all subsets of which are considered by TP. The runtime of PH also is very high rendering the approach infeasible for large datasets. We observed that the runtime of SM increased with the parameter $c$, which is as expected, since the product graph becomes more dense when the threshold parameter is raised. Therefore, we have also compared the runtime of PH and SM both using a Gaussian RBF kernel to compare distances, which leads to a very dense product graph. We found SM to be nevertheless approximately five times faster than PH, suggesting that our method of computation is superior in general. We also compared CSM to SM and observed that CSM is significantly faster on sparse graphs, while still reaching a comparable prediction accuracy.

## 5. Conclusion & Future Work

We have proposed a new graph kernel, which takes complex graph structures not containing repeated vertices into account and supports attributed graphs without restriction. The experimental evaluation shows promising results for attributed graphs from chem- and bioinformatics. Improving the runtime for large-scale datasets and large graphs remains future work. However, our approach already works well in practice for medium-sized graphs, large graphs when vertex and edge kernels are sparse, or when restricted to small or connected subgraphs. Thus, we believe subgraph matching kernels are a viable alternative to existing approaches for attributed graphs.

## Acknowledgments

We would like to thank Karsten Borgwardt and Nino Shervashidze for providing their kernel implementations and datasets. Nils Kriege was supported by the German Research Foundation (DFG), priority programme "Algorithm Engineering" (SPP 1307).

## References

Borgwardt, K.M. and Kriegel, H.-P. Shortest-path kernels on graphs. In *Proc. ICDM*, pp. 74–81, 2005.

Borgwardt, K.M., Ong, C.S., Schnauer, S., Vishwanathan, S.V.N., Smola, A.J., and Kriegel, H.-P. Protein function prediction via graph kernels. *Bioinformatics*, 21 Suppl 1:i47–i56, 2005.

---

[6]These distances were directly used for SP, which then reduces to a kernel basically comparing edges.

*Table 2.* Classification accuracies and runtimes in seconds for various kernels on graphs with simple labels. (Parameters used for runtime results: CSM $k = 5$, FLRW $l = 6$, TP $h = 3$, WL/WLSP $h = 3$; $^*$ – computation by explicit mapping)

| Method | Mutag | | MM | | FM | | MR | | FR | | Enzyme | |
|---|---|---|---|---|---|---|---|---|---|---|---|---|
| CSM | 85.4±1.2 | 30.5 | 63.3±1.7 | 29.2 | **63.8**±1.0 | 30.9 | 58.1±1.6 | 34.8 | 65.5±1.4 | 36.6 | 60.4±1.6 | 98m |
| TP | 86.7±0.9 | 10.0 | 66.1±1.1 | 14.0 | 60.5±1.3 | 14.9 | 56.5±2.2 | 16.0 | **68.0**±0.6 | 17.2 | 42.7±1.6 | 38m |
| FLRW | 87.1±0.7 | 8.3 | **66.6**±0.5 | 13.6 | 57.1±1.4 | 14.5 | 56.9±1.4 | 15.7 | 65.8±0.8 | 16.6 | 37.9±1.9 | 21m |
| GRW | **87.4**±1.1 | 40.1 | **66.4**±0.7 | 46.4 | 59.2±1.2 | 49.2 | 57.7±1.1 | 53.6 | **67.9**±0.5 | 56.7 | 31.6±1.3 | 229m |
| SP$^*$ | 83.3±1.4 | 0.2 | 61.0±2.1 | 0.15 | 60.4±1.7 | 0.18 | 56.2±2.7 | 0.17 | **67.9**±1.5 | 0.18 | 39.4±0.5 | 0.9 |
| WL$^*$ | 86.7±1.4 | 0.04 | 64.8±1.2 | 0.08 | 61.1±0.8 | 0.08 | 58.5±1.7 | 0.08 | 67.2±0.7 | 0.09 | 56.4±0.9 | 0.54 |
| WLSP$^*$ | 85.4±1.2 | 0.9 | **66.6**±1.9 | 1.16 | 60.4±1.3 | 1.32 | **59.7**±1.6 | 1.27 | 65.7±1.3 | 1.4 | **62.9**±1.0 | 25.7 |

*Table 3.* Classification accuracies and runtimes for various kernels on attributed graphs. (Parameters used for runtime results: (C)SM $k = 5$, FLRW $l = 6$, TP $h = 3$; Triangular Kernel $c = 0.25$)

| Method | Enzyme | | BZR | | COX2 | | DHFR | | ER | |
|---|---|---|---|---|---|---|---|---|---|---|
| (C)SM | **69.8**±0.7 | 115.5 | **79.4**±1.2 | 85.5 | **74.4**±1.7 | 195.5 | 79.9±1.1 | 116.2 | **82.0**±0.8 | 163.5 |
| PH | — | | 77.9±1.6 | 192m | **74.6**±1.5 | 364m | **80.8**±1.2 | 418m | 81.4±0.6 | 18h |
| FLRW | 63.9±0.3 | 56.2 | 77.9±1.1 | 54.7 | **74.4**±1.5 | 90.5 | 79.1±1.1 | 78.7 | 81.6±1.1 | 93.6 |
| SP | 65.7±1.1 | 282.2 | 78.2±1.2 | 40.6 | **74.5**±1.3 | 67.4 | 77.6±1.5 | 59.9 | 79.9±1.3 | 69.7 |
| TP | 55.5±0.6 | 69.8 | — | | — | | — | | — | |

Costa, F. and De Grave, K. Fast neighborhood subgraph pairwise distance kernel. In *Proc. ICML*, pp. 255–262, 2010.

Fröhlich, H., Wegner, J.K., Sieker, F., and Zell, A. Optimal assignment kernels for attributed molecular graphs. In *Proc. ICML*, pp. 225–232, 2005.

Gärtner, T., Flach, P., and Wrobel, S. On graph kernels: Hardness results and efficient alternatives. volume 2777 of *LNCS*, pp. 129–143. 2003.

Harchaoui, Z. and Bach, F. Image classification with segmentation graph kernels. In *Proc. CVPR*, 2007.

Haussler, D. Convolution kernels on discrete structures, 1999. Tech Rep UCSC-CRL-99-10.

Kashima, H., Tsuda, K., and Inokuchi, A. Marginalized kernels between labeled graphs. In *Proc. ICML*, pp. 321–328, 2003.

Koch, I. Enumerating all connected maximal common subgraphs in two graphs. *Theor. Comput. Sci.*, 250 (1-2):1–30, 2001.

Levi, G. A note on the derivation of maximal common subgraphs of two directed or undirected graphs. *Calcolo*, 1973.

Mahé, P. and Vert, J.-P. Graph kernels based on tree patterns for molecules. *Mach. Learn.*, 75:3–35, 2009.

Mahé, P., Ueda, N., Akutsu, T., Perret, J.-L., and Vert, J.-P. Extensions of marginalized graph kernels. In *Proc. ICML*, 2004.

Mahé, P., Ralaivola, L., Stoven, V., and Vert, J.-P. The pharmacophore kernel for virtual screening with support vector machines. *J Chem Inf Model*, 46(5): 2003–2014, 2006.

Menchetti, S., Costa, F., and Frasconi, P. Weighted decomposition kernels. In *Proc. ICML*, 2005.

Ramon, J. and Gärtner, T. Expressivity versus efficiency of graph kernels. In *First International Workshop on Mining Graphs, Trees and Sequences*, 2003.

Shervashidze, N., Vishwanathan, S.V.N., Petri, T.H., Mehlhorn, K., and Borgwardt, K.M. Efficient graphlet kernels for large graph comparison. In *AISTATS*, 2009.

Shervashidze, N., Schweitzer, P., van Leeuwen, E.J., Mehlhorn, K., and Borgwardt, K.M. Weisfeiler-lehman graph kernels. *JMLR*, 12:2539–2561, 2011.

Vishwanathan, S.V.N., Schraudolph, N.N., Kondor, R.I., and Borgwardt, K.M. Graph kernels. *JMLR*, 11:1201–1242, 2010.

Wale, N., Watson, I.A., and Karypis, G. Comparison of descriptor spaces for chemical compound retrieval and classification. *Knowl Inf Sys*, 14(3):347–375, 2008.